\documentclass[11pt,a4paper]{article}
\usepackage[hyperref]{emnlp2020}
\pdfoutput=1
\usepackage{times}
\usepackage{latexsym}

\usepackage{times}
\usepackage{latexsym}
\usepackage{amsmath}
\usepackage{url}
\usepackage{amssymb}
\usepackage{amsfonts}
\usepackage{graphicx}
\usepackage{tabularx}
\usepackage{multirow}
\usepackage{arydshln}
\usepackage{mathtools,nccmath}

\usepackage[utf8]{inputenc}
\usepackage[utf8]{vietnam}
\usepackage{enumitem}

\aclfinalcopy 

\title{PhoBERT: Pre-trained language models for Vietnamese}

\author{Dat Quoc Nguyen$^1$  \and Anh Tuan Nguyen$^{2,}$\thanks{\ \ Work done during internship at  VinAI Research.}  \\
  $^1$VinAI Research, Vietnam; $^2$NVIDIA, USA\\
   \tt{\normalsize v.datnq9@vinai.io, tuananhn@nvidia.com}}

\date{}

\begin{document}
\maketitle
\begin{abstract}
We present \textbf{PhoBERT} with two versions---PhoBERT\textsubscript{base} and PhoBERT\textsubscript{large}---the \emph{first} public large-scale monolingual language models pre-trained for Vietnamese. Experimental results show that PhoBERT consistently outperforms the recent best pre-trained multilingual  model XLM-R \citep{conneau2019unsupervised} and improves the state-of-the-art in multiple Vietnamese-specific NLP tasks including Part-of-speech tagging, Dependency parsing, Named-entity recognition and Natural language inference. We release PhoBERT  to facilitate future research and downstream applications for Vietnamese NLP. Our PhoBERT models are available at: \url{https://github.com/VinAIResearch/PhoBERT}.
\end{abstract}

\section{Introduction}\label{sec:intro}

Pre-trained language models, especially BERT \citep{devlin-etal-2019-bert}---the Bidirectional Encoder Representations from Transformers \citep{NIPS2017_7181}, have recently become  extremely popular and helped to produce significant improvement gains for various NLP tasks. The success of pre-trained BERT and its variants has largely been limited to the English language. For other languages, one could retrain a language-specific model using the BERT architecture \citep{abs-1906-08101,vries2019bertje,vu-xuan-etal-2019-etnlp,2019arXiv191103894M} or  employ existing pre-trained multilingual BERT-based models  \citep{devlin-etal-2019-bert,NIPS2019_8928,conneau2019unsupervised}.  

In terms of Vietnamese language modeling, to the best of our knowledge, there are two main concerns as follows: 

\begin{itemize}[leftmargin=*]
\setlength\itemsep{-1pt}
    \item The Vietnamese Wikipedia corpus is the only data used to train  monolingual language models \citep{vu-xuan-etal-2019-etnlp}, and it also is the only Vietnamese dataset which is included in the pre-training data used by all multilingual language models except XLM-R. It is worth noting that Wikipedia data is not representative of a general language use, and the Vietnamese Wikipedia data is relatively small (1GB in size uncompressed), while pre-trained language models can be significantly improved by using more pre-training data \cite{RoBERTa}. 
    
    \item All publicly released monolingual and multilingual BERT-based language models are not aware of the difference between Vietnamese syllables and word tokens. This ambiguity comes from the fact that the white space is also used to separate syllables that constitute words when written in Vietnamese.\footnote{\newcite{DinhQuangThang2008} show that 85\% of  Vietnamese word types are composed of at least two syllables.} 
    For example, a 6-syllable written text ``Tôi là một nghiên cứu viên'' (I am a researcher) forms 4 words ``Tôi\textsubscript{I} là\textsubscript{am} một\textsubscript{a} nghiên\_cứu\_viên\textsubscript{researcher}''. \\
Without doing a pre-process step of Vietnamese word segmentation, those models directly apply Byte-Pair encoding (BPE) methods \citep{sennrich-etal-2016-neural,kudo-richardson-2018-sentencepiece} to the syllable-level Vietnamese pre-training  data.\footnote{Although performing word segmentation before applying BPE on the Vietnamese Wikipedia corpus, ETNLP \citep{vu-xuan-etal-2019-etnlp} in fact {does not publicly release} any pre-trained BERT-based language model (\url{https://github.com/vietnlp/etnlp}). In particular, \newcite{vu-xuan-etal-2019-etnlp}  release a set of 15K BERT-based  word embeddings specialized only for the Vietnamese NER task.} 
Intuitively,  for word-level Vietnamese NLP tasks, those models pre-trained on  syllable-level data  might not perform as good as language models pre-trained on word-level data.

\end{itemize}

To handle the two concerns above, we train the {first} large-scale monolingual BERT-based ``base'' and ``large'' models  using a 20GB \textit{word-level} Vietnamese corpus. 
We evaluate our models on four downstream Vietnamese NLP tasks: the common word-level ones of Part-of-speech (POS) tagging,  Dependency parsing and Named-entity recognition (NER), and a language understanding task of Natural language inference (NLI) which can be formulated as either a syllable- or word-level task. Experimental results show that  our models obtain state-of-the-art (SOTA) results on all these  tasks. 
Our contributions are summarized as follows:

\begin{itemize}[leftmargin=*]
\setlength\itemsep{-1pt}
    \item We present the \textit{first} large-scale monolingual   language models pre-trained for Vietnamese.
    
    \item Our models help produce SOTA performances  on four downstream  tasks of POS tagging, Dependency parsing, NER and NLI, thus  showing  the effectiveness of large-scale BERT-based  monolingual language models for Vietnamese.
    
    \item To the best of our knowledge, we also perform the \textit{first} set of experiments to compare monolingual language models with the recent best multilingual model XLM-R in multiple (i.e. four) different language-specific tasks. The experiments show that our models outperform XLM-R   on all these  tasks, thus convincingly confirming that dedicated language-specific models still outperform multilingual ones.
    
    \item We publicly release our models under the name PhoBERT which can be used with  \texttt{fairseq}  \citep{ott2019fairseq} and  \texttt{transformers} \cite{Wolf2019HuggingFacesTS}. We hope that PhoBERT can serve as a strong baseline for future Vietnamese NLP research and  applications.
\end{itemize}

\section{PhoBERT} 

This section outlines the architecture and describes the   pre-training data and optimization setup that we use for PhoBERT.

\vspace{3pt}

\noindent\textbf{Architecture:}\  Our PhoBERT has two versions, PhoBERT\textsubscript{base} and PhoBERT\textsubscript{large}, using the same  architectures  of  BERT\textsubscript{base} and BERT\textsubscript{large}, respectively. PhoBERT pre-training approach is based on RoBERTa \citep{RoBERTa} which optimizes the BERT pre-training procedure for more robust performance. 

\vspace{3pt}

\noindent\textbf{Pre-training data:}\ To handle the first concern mentioned in Section  \ref{sec:intro}, we use a 20GB pre-training dataset of uncompressed texts. This dataset is a concatenation of two corpora: (i) the first one is the Vietnamese Wikipedia corpus ($\sim$1GB), and (ii) the second corpus ($\sim$19GB) is generated by removing similar articles and duplication from a 50GB Vietnamese news corpus.\footnote{\url{https://github.com/binhvq/news-corpus}, crawled from a wide range of news websites  and  topics.} To solve the second concern,  
we employ RDRSegmenter \citep{nguyen-etal-2018-fast} from VnCoreNLP \citep{vu-etal-2018-vncorenlp} to perform word and sentence segmentation on the pre-training dataset, resulting in $\sim$145M word-segmented sentences  ($\sim$3B word tokens). Different from RoBERTa, we then apply \texttt{fastBPE} \citep{sennrich-etal-2016-neural} to segment these sentences with subword units, using a vocabulary of 64K subword types. On average there are 24.4 subword tokens per sentence. 

\vspace{3pt}

\noindent\textbf{Optimization:}\  We employ the RoBERTa implementation in  \texttt{fairseq}  \citep{ott2019fairseq}. We set a maximum length at 256 subword tokens, thus generating 145M $\times$ 24.4 / 256 $\approx$ 13.8M sentence blocks. Following \newcite{RoBERTa}, we optimize the models using Adam \citep{KingmaB14}.  We use a batch size of 1024 across 4 V100 GPUs (16GB each) and a peak learning rate of 0.0004 for PhoBERT\textsubscript{base}, and a batch size of 512 and a peak learning rate of 0.0002 for PhoBERT\textsubscript{large}. We run for 40 epochs (here, the learning rate is warmed up for 2 epochs), thus resulting in 13.8M $\times$ 40 / 1024 $\approx$ 540K training steps for PhoBERT\textsubscript{base} and 1.08M training steps for PhoBERT\textsubscript{large}. We pre-train PhoBERT\textsubscript{base} during  3 weeks, and then PhoBERT\textsubscript{large} during  5 weeks.

\begin{table}[!t]
    \centering
    \begin{tabular}{l|l|l|l}
    \hline
    \textbf{Task}  & \textbf{\#training} & \textbf{\#valid} & \textbf{\#test} \\
    \hline
    
    POS tagging$^\dagger$ & 27,000 & 870 & 2,120 \\
    Dep. parsing$^\dagger$ & 8,977 & 200 & 1,020 \\
    NER$^\dagger$ & 14,861 & 2,000 & 2,831\\
    NLI$^\ddagger$ & 392,702 & 2,490 & 5,010\\
    \hline
    \end{tabular}
    \caption{Statistics of the downstream task datasets. ``\#training'', ``\#valid''  and  ``\#test'' denote the size of the training, validation and test sets, respectively. $\dagger$ and $\ddagger$ refer to the dataset size as   the numbers of sentences and sentence pairs, respectively.}
    \label{tab:data}
\end{table}

  \begin{table*}[!ht]
     \centering
      \resizebox{15.5cm}{!}{
     \begin{tabular}{l|l|l|l}
    \hline
          \multicolumn{2}{c|}{\textbf{POS tagging} (word-level)} & \multicolumn{2}{c}{\textbf{Dependency parsing} (word-level)}\\
    \hline
    Model & Acc. & Model & LAS / UAS \\
    \hline 
    RDRPOSTagger \citep{nguyen-etal-2014-rdrpostagger} [$\clubsuit$] &  95.1 & \_ & \_  \\
    
    BiLSTM-CNN-CRF \citep{ma-hovy-2016-end} [$\clubsuit$] & 95.4 & VnCoreNLP-DEP \citep{vu-etal-2018-vncorenlp} [$\bigstar$]  & 71.38 / 77.35 \\

    VnCoreNLP-POS  \citep{nguyen-etal-2017-word} [$\clubsuit$] & 95.9 &jPTDP-v2  [$\bigstar$] & 73.12 / 79.63 \\
    
   jPTDP-v2 \citep{nguyen-verspoor-2018-improved} [$\bigstar$] & 95.7  &jointWPD [$\bigstar$]  & 73.90 / 80.12  \\
    
    jointWPD \citep{nguyen-2019-neural} [$\bigstar$]  & 96.0 & Biaffine \citep{DozatM17} [$\bigstar$]  & 74.99 / 81.19   \\
    
    XLM-R\textsubscript{base} (our result) & 96.2  & Biaffine w/ XLM-R\textsubscript{base} (our result) &  76.46 /  83.10  \\
    
    XLM-R\textsubscript{large} (our result) & 96.3 & Biaffine w/ XLM-R\textsubscript{large} (our result) & 75.87 / 82.70   \\
    
    \hline
    PhoBERT\textsubscript{base} & \underline{96.7} & Biaffine w/ PhoBERT\textsubscript{base} & \textbf{78.77} / \textbf{85.22}  \\
    
    PhoBERT\textsubscript{large} & \textbf{96.8} & Biaffine w/ PhoBERT\textsubscript{large} & \underline{77.85} / \underline{84.32}   \\
    \hline 
     \end{tabular}
     }
     \caption{Performance scores (in \%) on the POS tagging and Dependency parsing test sets. ``Acc.'', ``LAS'' and ``UAS'' abbreviate the Accuracy, the Labeled Attachment Score and the Unlabeled Attachment Score, respectively (here, all these evaluation metrics are computed on all word tokens, including punctuation).
     [$\clubsuit$] and [$\bigstar$] denote 
    results  reported by  \newcite{nguyen-etal-2017-word} and  \newcite{nguyen-2019-neural}, respectively.}
     \label{tab:posdep}
 \end{table*}

\section{Experimental setup}

 We evaluate the performance of PhoBERT on four  downstream Vietnamese NLP tasks: POS tagging, Dependency parsing, NER and NLI.

\subsubsection*{Downstream task datasets} 

Table \ref{tab:data} presents the statistics of the experimental datasets that we employ for downstream task evaluation. 
For POS tagging, Dependency parsing  and NER, we follow the  VnCoreNLP setup   \citep{vu-etal-2018-vncorenlp}, using standard benchmarks of the VLSP 2013 POS tagging dataset,\footnote{\url{https://vlsp.org.vn/vlsp2013/eval}} the VnDT dependency treebank v1.1 \cite{Nguyen2014NLDB} with   POS tags predicted by VnCoreNLP and the VLSP 2016 NER dataset \citep{JCC13161}. 

For NLI, we use the manually-constructed Vietnamese validation and test sets  from the cross-lingual NLI (XNLI) corpus v1.0 \citep{conneau-etal-2018-xnli} where the Vietnamese  training set is released  as a machine-translated version of the corresponding English training set \citep{N18-1101}. 
Unlike the  POS tagging, Dependency parsing   and NER datasets which provide the gold word segmentation, for NLI, we employ RDRSegmenter to segment the text into words before applying BPE to produce subwords from word tokens. 
 
\subsubsection*{Fine-tuning}

Following \newcite{devlin-etal-2019-bert}, for POS tagging and NER, we append a linear prediction layer on top of the PhoBERT architecture (i.e. to the last Transformer layer of PhoBERT) w.r.t. the first  subword  of each word token.\footnote{In our preliminary experiments, using the average of contextualized embeddings of subword tokens of each word to represent the word produces slightly lower performance than using the contextualized embedding of the first subword.}  
For dependency parsing, following \newcite{nguyen-2019-neural}, we employ a reimplementation of the state-of-the-art Biaffine dependency  parser \citep{DozatM17} from \newcite{ma-etal-2018-stack} with default optimal hyper-parameters. 
We then extend this parser  by replacing the pre-trained word embedding of each word in an input sentence by the corresponding contextualized embedding (from the last layer) computed for the first subword token of the word. 

For POS tagging, NER and NLI, we employ \texttt{transformers} \cite{Wolf2019HuggingFacesTS} to fine-tune PhoBERT for each task and each dataset independently. We use AdamW \citep{loshchilov2018decoupled} with a fixed learning rate of 1.e-5 and a batch size of 32 \citep{RoBERTa}. We fine-tune in 30 training epochs, evaluate the task performance after each epoch on the validation set  (here, early stopping is applied when there is no improvement after 5 continuous epochs), and then select the best model checkpoint to report the final result on the test set (note that each of our scores is an average over 5 runs with different random seeds). 

  \begin{table*}[!ht]
     \centering
     \resizebox{15.5cm}{!}{
     \begin{tabular}{l|l|l|l}
    \hline
          \multicolumn{2}{c|}{\textbf{NER} (word-level)} & \multicolumn{2}{c}{\textbf{NLI} (syllable- or word-level)} \\
          
    \hline
    Model & F\textsubscript{1} &  Model & Acc. \\
    \hline 
    BiLSTM-CNN-CRF [$\blacklozenge$]  & 88.3 & \_ & \_\\
    
    VnCoreNLP-NER \citep{vu-etal-2018-vncorenlp} [$\blacklozenge$] & 88.6  & BiLSTM-max \citep{conneau-etal-2018-xnli} & 66.4  \\

    VNER \citep{8713740} & 89.6 &  mBiLSTM \citep{ArtetxeS19} & 72.0  \\
    
    BiLSTM-CNN-CRF + ETNLP [$\spadesuit$] & 91.1  & multilingual BERT \citep{devlin-etal-2019-bert} [$\blacksquare$]  & 69.5  \\
    
    VnCoreNLP-NER + ETNLP [$\spadesuit$] & 91.3   &  XLM\textsubscript{MLM+TLM} \citep{NIPS2019_8928} & 76.6  \\
    
    XLM-R\textsubscript{base} (our result) & 92.0  & XLM-R\textsubscript{base} \citep{conneau2019unsupervised} & {75.4} \\
     
    XLM-R\textsubscript{large} (our result) & 92.8   &  XLM-R\textsubscript{large} \citep{conneau2019unsupervised} & \underline{79.7} \\
    
    \hline
    PhoBERT\textsubscript{base}& \underline{93.6} & PhoBERT\textsubscript{base}& {78.5} \\
    
    PhoBERT\textsubscript{large}& \textbf{94.7}   & PhoBERT\textsubscript{large}& \textbf{80.0} \\
    \hline

     \end{tabular}
    }
     \caption{Performance scores (in \%) on  the NER and NLI test sets.
      [$\blacklozenge$], [$\spadesuit$] and [$\blacksquare$] denote 
    results  reported by  \newcite{vu-etal-2018-vncorenlp},  \newcite{vu-xuan-etal-2019-etnlp} and \newcite{wu-dredze-2019-beto}, respectively.  
    Note that there are higher Vietnamese NLI results reported  for XLM-R when fine-tuning on the concatenation of all 15  training datasets from the XNLI corpus (i.e. TRANSLATE-TRAIN-ALL: 79.5\% for XLM-R\textsubscript{base} and 83.4\% XLM-R\textsubscript{large}). However, those results might not be comparable  as we only use the  monolingual Vietnamese  training data for fine-tuning. }
     \label{tab:nernli}
 \end{table*}
 
\section{Experimental results}\label{sec:results}
 
\subsubsection*{Main results} 

Tables \ref{tab:posdep} and \ref{tab:nernli} compare  PhoBERT scores with the previous highest reported results, using the same experimental setup. It is clear that our PhoBERT helps produce new SOTA performance results  for all four downstream tasks. 
 
For  \underline{POS tagging}, the neural model jointWPD for joint POS tagging and dependency parsing \citep{nguyen-2019-neural} and the feature-based model VnCoreNLP-POS \citep{nguyen-etal-2017-word} are the two previous SOTA models, obtaining accuracies at  about 96.0\%.  PhoBERT obtains 0.8\% absolute higher accuracy than these two models. 

For \underline{Dependency parsing}, the previous highest parsing scores LAS and UAS are obtained by the Biaffine  parser at 75.0\% and 81.2\%, respectively.  PhoBERT helps boost the Biaffine parser with about 4\% absolute improvement, achieving a LAS at 78.8\% and a UAS at 85.2\%.

For  \underline{NER}, PhoBERT\textsubscript{large} produces 1.1 points higher F\textsubscript{1} than PhoBERT\textsubscript{base}. In addition,  PhoBERT\textsubscript{base} obtains 2+ points higher than the previous SOTA feature- and neural network-based models VnCoreNLP-NER \citep{vu-etal-2018-vncorenlp} and BiLSTM-CNN-CRF \citep{ma-hovy-2016-end} which are trained with the set of 15K BERT-based ETNLP word embeddings  \citep{vu-xuan-etal-2019-etnlp}. 
 
 For  \underline{NLI}, 
PhoBERT outperforms the multilingual BERT \citep{devlin-etal-2019-bert}  and the BERT-based cross-lingual model with a new translation language modeling objective XLM\textsubscript{MLM+TLM} \citep{NIPS2019_8928}  by large margins.   PhoBERT also performs  better than the recent best pre-trained multilingual model XLM-R   but  using far fewer parameters than XLM-R:  135M (PhoBERT\textsubscript{base}) vs.  250M (XLM-R\textsubscript{base});  370M (PhoBERT\textsubscript{large}) vs.  560M (XLM-R\textsubscript{large}).

\subsubsection*{Discussion}

We find that PhoBERT\textsubscript{large} achieves 0.9\% lower dependency parsing scores than  PhoBERT\textsubscript{base}. One possible reason is that the last Transformer layer in the BERT architecture might not be the optimal one which encodes the richest information of syntactic structures \cite{hewitt-manning-2019-structural,jawahar-etal-2019-bert}.  Future work will study which PhoBERT's Transformer layer contains richer syntactic information by evaluating the Vietnamese parsing performance from each layer.

Using more pre-training data can significantly improve the quality of the pre-trained language models \cite{RoBERTa}. Thus it is not surprising that PhoBERT helps produce better performance than  ETNLP on NER, and the multilingual BERT and XLM\textsubscript{MLM+TLM} on NLI (here, PhoBERT uses 20GB of Vietnamese texts while those models employ the 1GB Vietnamese Wikipedia corpus).  

Following the fine-tuning approach that we use for PhoBERT, we carefully fine-tune XLM-R for the remaining Vietnamese POS tagging, Dependency parsing and NER tasks (here, it is  applied to the first sub-syllable token of the first syllable of each word).\footnote{For fine-tuning XLM-R, we use a grid search on the validation set to select the AdamW learning rate from \{5e-6, 1e-5, 2e-5, 4e-5\} and the batch size from \{16, 32\}.} 
Tables \ref{tab:posdep} and \ref{tab:nernli} show  that our PhoBERT also does better than   XLM-R on these three word-level tasks. 
It is worth noting that XLM-R uses a 2.5TB pre-training corpus which contains 137GB of  Vietnamese texts (i.e. about 137\ /\ 20 $\approx$ 7 times bigger than our  pre-training corpus). 
Recall that PhoBERT performs Vietnamese word segmentation to segment  syllable-level  sentences  into word tokens before applying BPE to segment the word-segmented sentences into subword units, while XLM-R directly applies BPE to the syllable-level Vietnamese pre-training  sentences. 
 This  reconfirms that the dedicated language-specific models still outperform the  multilingual ones \citep{2019arXiv191103894M}.\footnote{Note that \newcite{2019arXiv191103894M} only  compare their model CamemBERT with XLM-R on the French NLI task.}

 \section{Conclusion}
 
In this paper, we have presented the first large-scale monolingual  PhoBERT language models pre-trained for Vietnamese. We demonstrate the usefulness of PhoBERT by showing that  PhoBERT  performs better than the recent best multilingual model XLM-R and helps produce the SOTA performances for four downstream Vietnamese NLP tasks of POS tagging, Dependency parsing, NER and NLI. 
By publicly releasing PhoBERT models, 
we hope that they can foster future research and applications in Vietnamese NLP. 
 
{
\bibliographystyle{acl_natbib}
\bibliography{REFs}

\begin{thebibliography}{34}
\expandafter\ifx\csname natexlab\endcsname\relax\def\natexlab#1{#1}\fi

\bibitem[{Artetxe and Schwenk(2019)}]{ArtetxeS19}
Mikel Artetxe and Holger Schwenk. 2019.
\newblock {Massively Multilingual Sentence Embeddings for Zero-Shot
  Cross-Lingual Transfer and Beyond}.
\newblock \emph{{TACL}}, 7:597--610.

\bibitem[{Conneau et~al.(2020)Conneau, Khandelwal, Goyal, Chaudhary, Wenzek,
  Guzm{\'a}n, Grave, Ott, Zettlemoyer, and Stoyanov}]{conneau2019unsupervised}
Alexis Conneau, Kartikay Khandelwal, Naman Goyal, Vishrav Chaudhary, Guillaume
  Wenzek, Francisco Guzm{\'a}n, Edouard Grave, Myle Ott, Luke Zettlemoyer, and
  Veselin Stoyanov. 2020.
\newblock \href {https://arxiv.org/pdf/1911.02116v1.pdf} {{Unsupervised
  Cross-lingual Representation Learning at Scale}}.
\newblock In \emph{Proceedings of ACL}, pages 8440--8451.

\bibitem[{Conneau and Lample(2019)}]{NIPS2019_8928}
Alexis Conneau and Guillaume Lample. 2019.
\newblock {Cross-lingual Language Model Pretraining}.
\newblock In \emph{Proceedings of NeurIPS}, pages 7059--7069.

\bibitem[{Conneau et~al.(2018)Conneau, Rinott, Lample, Schwenk, Stoyanov,
  Williams, and Bowman}]{conneau-etal-2018-xnli}
Alexis Conneau, Ruty Rinott, Guillaume Lample, Holger Schwenk, Ves Stoyanov,
  Adina Williams, and Samuel~R. Bowman. 2018.
\newblock {XNLI}: Evaluating cross-lingual sentence representations.
\newblock In \emph{Proceedings of EMNLP}, pages 2475--2485.

\bibitem[{Cui et~al.(2019)Cui, Che, Liu, Qin, Yang, Wang, and
  Hu}]{abs-1906-08101}
Yiming Cui, Wanxiang Che, Ting Liu, Bing Qin, Ziqing Yang, Shijin Wang, and
  Guoping Hu. 2019.
\newblock {Pre-Training with Whole Word Masking for Chinese BERT}.
\newblock \emph{arXiv preprint}, arXiv:1906.08101.

\bibitem[{Devlin et~al.(2019)Devlin, Chang, Lee, and
  Toutanova}]{devlin-etal-2019-bert}
Jacob Devlin, Ming-Wei Chang, Kenton Lee, and Kristina Toutanova. 2019.
\newblock {BERT}: Pre-training of deep bidirectional transformers for language
  understanding.
\newblock In \emph{Proceedings of NAACL}, pages 4171--4186.

\bibitem[{Dozat and Manning(2017)}]{DozatM17}
Timothy Dozat and Christopher~D. Manning. 2017.
\newblock {Deep Biaffine Attention for Neural Dependency Parsing}.
\newblock In \emph{Proceedings of ICLR}.

\bibitem[{Hewitt and Manning(2019)}]{hewitt-manning-2019-structural}
John Hewitt and Christopher~D. Manning. 2019.
\newblock {A} structural probe for finding syntax in word representations.
\newblock In \emph{Proceedings of NAACL}, pages 4129--4138.

\bibitem[{Jawahar et~al.(2019)Jawahar, Sagot, and
  Seddah}]{jawahar-etal-2019-bert}
Ganesh Jawahar, Beno{\^\i}t Sagot, and Djam{\'e} Seddah. 2019.
\newblock What does {BERT} learn about the structure of language?
\newblock In \emph{Proceedings of ACL}, pages 3651--3657.

\bibitem[{Kingma and Ba(2014)}]{KingmaB14}
Diederik~P. Kingma and Jimmy Ba. 2014.
\newblock {Adam: {A} Method for Stochastic Optimization}.
\newblock \emph{arXiv preprint}, arXiv:1412.6980.

\bibitem[{Kudo and Richardson(2018)}]{kudo-richardson-2018-sentencepiece}
Taku Kudo and John Richardson. 2018.
\newblock {{S}entence{P}iece: A simple and language independent subword
  tokenizer and detokenizer for Neural Text Processing}.
\newblock In \emph{Proceedings of EMNLP: System Demonstrations}, pages 66--71.

\bibitem[{Liu et~al.(2019)Liu, Ott, Goyal, Du, Joshi, Chen, Levy, Lewis,
  Zettlemoyer, and Stoyanov}]{RoBERTa}
Yinhan Liu, Myle Ott, Naman Goyal, Jingfei Du, Mandar Joshi, Danqi Chen, Omer
  Levy, Mike Lewis, Luke Zettlemoyer, and Veselin Stoyanov. 2019.
\newblock {RoBERTa: {A} Robustly Optimized {BERT} Pretraining Approach}.
\newblock \emph{arXiv preprint}, arXiv:1907.11692.

\bibitem[{Loshchilov and Hutter(2019)}]{loshchilov2018decoupled}
Ilya Loshchilov and Frank Hutter. 2019.
\newblock {Decoupled Weight Decay Regularization}.
\newblock In \emph{Proceedings of ICLR}.

\bibitem[{Ma and Hovy(2016)}]{ma-hovy-2016-end}
Xuezhe Ma and Eduard Hovy. 2016.
\newblock End-to-end sequence labeling via bi-directional {LSTM}-{CNN}s-{CRF}.
\newblock In \emph{Proceedings of ACL}, pages 1064--1074.

\bibitem[{Ma et~al.(2018)Ma, Hu, Liu, Peng, Neubig, and
  Hovy}]{ma-etal-2018-stack}
Xuezhe Ma, Zecong Hu, Jingzhou Liu, Nanyun Peng, Graham Neubig, and Eduard
  Hovy. 2018.
\newblock {Stack-Pointer Networks for Dependency Parsing}.
\newblock In \emph{Proceedings of ACL}, pages 1403--1414.

\bibitem[{{Martin} et~al.(2020){Martin}, {Muller}, {Ortiz Su{\'a}rez},
  {Dupont}, {Romary}, {Villemonte de la Clergerie}, {Seddah}, and
  {Sagot}}]{2019arXiv191103894M}
Louis {Martin}, Benjamin {Muller}, Pedro~Javier {Ortiz Su{\'a}rez}, Yoann
  {Dupont}, Laurent {Romary}, {\'E}ric {Villemonte de la Clergerie}, Djam{\'e}
  {Seddah}, and Beno{\^\i}t {Sagot}. 2020.
\newblock {CamemBERT: a Tasty French Language Model}.
\newblock In \emph{Proceedings of ACL}, pages 7203--7219.

\bibitem[{Nguyen(2019)}]{nguyen-2019-neural}
Dat~Quoc Nguyen. 2019.
\newblock A neural joint model for {V}ietnamese word segmentation, {POS}
  tagging and dependency parsing.
\newblock In \emph{Proceedings of ALTA}, pages 28--34.

\bibitem[{Nguyen et~al.(2014{\natexlab{a}})Nguyen, Nguyen, Pham, and
  Pham}]{nguyen-etal-2014-rdrpostagger}
Dat~Quoc Nguyen, Dai~Quoc Nguyen, Dang~Duc Pham, and Son~Bao Pham.
  2014{\natexlab{a}}.
\newblock {RDRPOSTagger: A Ripple Down Rules-based Part-Of-Speech Tagger}.
\newblock In \emph{Proceedings of the Demonstrations at EACL}, pages 17--20.

\bibitem[{Nguyen et~al.(2014{\natexlab{b}})Nguyen, Nguyen, Pham, Nguyen, and
  Nguyen}]{Nguyen2014NLDB}
Dat~Quoc Nguyen, Dai~Quoc Nguyen, Son~Bao Pham, Phuong-Thai Nguyen, and Minh~Le
  Nguyen. 2014{\natexlab{b}}.
\newblock {From Treebank Conversion to Automatic Dependency Parsing for
  Vietnamese}.
\newblock In \emph{{Proceedings of NLDB}}, pages 196--207.

\bibitem[{Nguyen et~al.(2018)Nguyen, Nguyen, Vu, Dras, and
  Johnson}]{nguyen-etal-2018-fast}
Dat~Quoc Nguyen, Dai~Quoc Nguyen, Thanh Vu, Mark Dras, and Mark Johnson. 2018.
\newblock {A Fast and Accurate Vietnamese Word Segmenter}.
\newblock In \emph{Proceedings of LREC}, pages 2582--2587.

\bibitem[{Nguyen and Verspoor(2018)}]{nguyen-verspoor-2018-improved}
Dat~Quoc Nguyen and Karin Verspoor. 2018.
\newblock An improved neural network model for joint {POS} tagging and
  dependency parsing.
\newblock In \emph{Proceedings of the {C}o{NLL} 2018 Shared Task}, pages
  81--91.

\bibitem[{Nguyen et~al.(2017)Nguyen, Vu, Nguyen, Dras, and
  Johnson}]{nguyen-etal-2017-word}
Dat~Quoc Nguyen, Thanh Vu, Dai~Quoc Nguyen, Mark Dras, and Mark Johnson. 2017.
\newblock From word segmentation to {POS} tagging for {V}ietnamese.
\newblock In \emph{Proceedings of ALTA}, pages 108--113.

\bibitem[{Nguyen et~al.(2019{\natexlab{a}})Nguyen, Ngo, Vu, Tran, and
  Nguyen}]{JCC13161}
Huyen Nguyen, Quyen Ngo, Luong Vu, Vu~Tran, and Hien Nguyen.
  2019{\natexlab{a}}.
\newblock {VLSP Shared Task: Named Entity Recognition}.
\newblock \emph{Journal of Computer Science and Cybernetics}, 34(4):283--294.

\bibitem[{Nguyen et~al.(2019{\natexlab{b}})Nguyen, Dong, and Nguyen}]{8713740}
Kim~Anh Nguyen, Ngan Dong, and Cam-Tu Nguyen. 2019{\natexlab{b}}.
\newblock {Attentive Neural Network for Named Entity Recognition in
  Vietnamese}.
\newblock In \emph{Proceedings of RIVF}.

\bibitem[{Ott et~al.(2019)Ott, Edunov, Baevski, Fan, Gross, Ng, Grangier, and
  Auli}]{ott2019fairseq}
Myle Ott, Sergey Edunov, Alexei Baevski, Angela Fan, Sam Gross, Nathan Ng,
  David Grangier, and Michael Auli. 2019.
\newblock {fairseq: A Fast, Extensible Toolkit for Sequence Modeling}.
\newblock In \emph{Proceedings of NAACL-HLT 2019: Demonstrations}, pages
  48--53.

\bibitem[{Sennrich et~al.(2016)Sennrich, Haddow, and
  Birch}]{sennrich-etal-2016-neural}
Rico Sennrich, Barry Haddow, and Alexandra Birch. 2016.
\newblock {Neural Machine Translation of Rare Words with Subword Units}.
\newblock In \emph{Proceedings of ACL}, pages 1715--1725.

\bibitem[{Thang et~al.(2008)Thang, Phuong, Huyen, Tu, Rossignol, and
  Luong}]{DinhQuangThang2008}
Dinh~Quang Thang, Le~Hong Phuong, Nguyen Thi~Minh Huyen, Nguyen~Cam Tu, Mathias
  Rossignol, and Vu~Xuan Luong. 2008.
\newblock {Word segmentation of Vietnamese texts: a comparison of approaches}.
\newblock In \emph{Proceedings of LREC}, pages 1933--1936.

\bibitem[{Vaswani et~al.(2017)Vaswani, Shazeer, Parmar, Uszkoreit, Jones,
  Gomez, Kaiser, and Polosukhin}]{NIPS2017_7181}
Ashish Vaswani, Noam Shazeer, Niki Parmar, Jakob Uszkoreit, Llion Jones,
  Aidan~N Gomez, {\L}ukasz Kaiser, and Illia Polosukhin. 2017.
\newblock {Attention is All you Need}.
\newblock In \emph{Advances in Neural Information Processing Systems 30}, pages
  5998--6008.

\bibitem[{de~Vries et~al.(2019)de~Vries, van Cranenburgh, Bisazza, Caselli, van
  Noord, and Nissim}]{vries2019bertje}
Wietse de~Vries, Andreas van Cranenburgh, Arianna Bisazza, Tommaso Caselli,
  Gertjan van Noord, and Malvina Nissim. 2019.
\newblock {BERTje: A Dutch BERT Model}.
\newblock \emph{arXiv preprint}, arXiv:1912.09582.

\bibitem[{Vu et~al.(2018)Vu, Nguyen, Nguyen, Dras, and
  Johnson}]{vu-etal-2018-vncorenlp}
Thanh Vu, Dat~Quoc Nguyen, Dai~Quoc Nguyen, Mark Dras, and Mark Johnson. 2018.
\newblock {VnCoreNLP: A Vietnamese Natural Language Processing Toolkit}.
\newblock In \emph{Proceedings of NAACL: Demonstrations}, pages 56--60.

\bibitem[{Vu et~al.(2019)Vu, Vu, Tran, and Jiang}]{vu-xuan-etal-2019-etnlp}
Xuan-Son Vu, Thanh Vu, Son Tran, and Lili Jiang. 2019.
\newblock {ETNLP}: A visual-aided systematic approach to select pre-trained
  embeddings for a downstream task.
\newblock In \emph{Proceedings of RANLP}, pages 1285--1294.

\bibitem[{Williams et~al.(2018)Williams, Nangia, and Bowman}]{N18-1101}
Adina Williams, Nikita Nangia, and Samuel Bowman. 2018.
\newblock {A Broad-Coverage Challenge Corpus for Sentence Understanding through
  Inference}.
\newblock In \emph{Proceedings of NAACL}, pages 1112--1122.

\bibitem[{Wolf et~al.(2019)Wolf, Debut, Sanh, Chaumond, Delangue, Moi, Cistac,
  Rault, Louf, Funtowicz, and Brew}]{Wolf2019HuggingFacesTS}
Thomas Wolf, Lysandre Debut, Victor Sanh, Julien Chaumond, Clement Delangue,
  Anthony Moi, Pierric Cistac, Tim Rault, R'emi Louf, Morgan Funtowicz, and
  Jamie Brew. 2019.
\newblock {HuggingFace's Transformers: State-of-the-art Natural Language
  Processing}.
\newblock \emph{arXiv preprint}, arXiv:1910.03771.

\bibitem[{Wu and Dredze(2019)}]{wu-dredze-2019-beto}
Shijie Wu and Mark Dredze. 2019.
\newblock Beto, bentz, becas: The surprising cross-lingual effectiveness of
  {BERT}.
\newblock In \emph{Proceedings of EMNLP-IJCNLP}, pages 833--844.

\end{thebibliography}
}

\end{document}